\documentclass{article} % For LaTeX2e
\usepackage{iclr2027_conference,times}
\usepackage{svg}
\usepackage{centernot}

\usepackage[utf8]{inputenc} % allow utf-8 input
\usepackage[T1]{fontenc}    % use 8-bit T1 fonts
\usepackage[backref=page]{hyperref}       % hyperlinks
\usepackage{url}            % simple URL typesetting
\usepackage{booktabs}       % professional-quality tables
\usepackage{threeparttable}
\usepackage{amsfonts}       % blackboard math symbols
\usepackage{nicefrac}       % compact symbols for 1/2, etc.
\usepackage{microtype}      % microtypography
\usepackage{xcolor}         % colors

\usepackage[utf8]{inputenc} % allow utf-8 input
\usepackage[T1]{fontenc}    % use 8-bit T1 fonts
\usepackage{hyperref}       % hyperlinks
\usepackage{url}            % simple URL typesetting
\usepackage{booktabs}       % professional-quality tables
\usepackage{enumitem}
\usepackage{mathtools}
\usepackage{todonotes}
\usepackage{amssymb,fge}
\usepackage{thm-restate}
\usepackage{wrapfig}
\usepackage{bbm}
\usepackage{mathrsfs}
\usepackage{soul}
\usepackage{array}
\usepackage{multirow}
\usepackage{algorithm}
\usepackage{tikz}
\usepackage{subcaption}
\usepackage{wrapfig}
\usetikzlibrary{positioning,arrows.meta,shapes.geometric}

\usepackage{algorithm}
\usepackage{listings}
\usepackage{nicefrac}       % compact symbols for 1/2, etc.
\usepackage{microtype}      % microtypography
\usepackage{xcolor}         % colors
\usepackage{colortbl}
\usepackage{enumitem}
\usepackage{xspace}
\usepackage{lineno}
\usepackage{titletoc}
\usepackage{amsmath,amssymb,booktabs,array,tabularx}
\PassOptionsToPackage{final,pagebackref=true}{hyperref}
\usepackage{hyperref} % adds hyper links inside the generated pdf file
\usepackage{cleveref}
\usepackage{wasysym}
\usepackage{acronym}
\usepackage{xfrac}

\usepackage{amsmath,amsfonts,bm}

\def\eqref#1{equation~\ref{#1}}
\def\1{\bm{1}}

\DeclareMathAlphabet{\mathsfit}{\encodingdefault}{\sfdefault}{m}{sl}
\SetMathAlphabet{\mathsfit}{bold}{\encodingdefault}{\sfdefault}{bx}{n}

\def\gA{{\mathcal{A}}}

\def\gJ{{\mathcal{J}}}

\def\gL{{\mathcal{L}}}
\def\gM{{\mathcal{M}}}
\def\gN{{\mathcal{N}}}

\def\gP{{\mathcal{P}}}

\def\gS{{\mathcal{S}}}

\def\gU{{\mathcal{U}}}

\newcommand{\E}{\mathbb{E}}

\newcommand{\R}{\mathbb{R}}

\usepackage{enumitem}
\usepackage[framemethod=TikZ]{mdframed}
\newcommand{\xhdr}[1]{{\noindent\bfseries #1}.}
\mdfdefinestyle{MyFrame2}{%
    linecolor=white,
    outerlinewidth=1pt,
    roundcorner=1pt,
    innertopmargin=3pt,
    innerbottommargin=2pt,
    innerrightmargin=7pt,
    innerleftmargin=7pt,
    backgroundcolor=black!3!white}

\newcommand{\nameshort}{\textsc{VLA-SSD}\xspace}

\usepackage{amsmath, amsthm, mathtools}
\usepackage{cleveref}
\usepackage[normalem]{ulem}
\usepackage{changepage}
\usepackage{float}
\usepackage{needspace} % preamble
\usepackage{makecell}
\usepackage{pifont}
\usepackage{tikz}

\definecolor{sectiongray}{gray}{0.90}
\definecolor{refgray}{gray}{0.55}
\usepackage{algorithm}
\usepackage{algorithmicx}
\usepackage{algcompatible}
\usepackage{algpseudocode}

\usepackage{amsthm}
\newtheorem{theorem}{Theorem}
\newtheorem{proposition}[theorem]{Proposition}
\usepackage{graphicx}
\usepackage{booktabs}
\usepackage{capt-of}
\usepackage{graphicx}
\usepackage{booktabs}
\usepackage{capt-of}
\newsavebox{\diversitybox}
\newlength{\diversitytrim}

\hypersetup{
	colorlinks=true,       % false: boxed links; true: colored links
	linkcolor=blue,        % color of internal links
	citecolor=blue,        % color of links to bibliography
	filecolor=magenta,     % color of file links
	urlcolor=blue         
}

\title{Stochastic Self-Distillation for Flow-Based Vision-Language-Action Models}

\author{Jingru Guo \quad Alessandra Russo \quad Joey Bose \quad Christos Ziakas  \\[4pt]
Imperial College London
}

\iclrfinalcopy
\begin{document}

\maketitle
\lhead{}
\makeatletter
\begingroup
\renewcommand{\thefootnote}{}
\long\def\@makefntext#1{\noindent#1}
\footnotetext{%
Email: \texttt{jingru.guo25@imperial.ac.uk, c.ziakas24@imperial.ac.uk}\\
Code: \url{https://github.com/llexieguo/VLA-SSD.git}%
}
\endgroup
\makeatother

\looseness=-1

\begin{abstract}
\looseness=-1
Flow-based vision--language--action (VLA) models achieve strong performance in robotics through imitation learning, yet fine-tuning them with on-policy reinforcement learning (RL) is limited. In group-based RL methods like Group Relative Policy Optimization (GRPO), every rollout step requires multi-step ODE integration, and the sampler's limited stochasticity leads to rollout groups with low variance. We propose VLA-SSD, which addresses both high inference latency and limited stochasticity. In particular, our method self-distills a pretrained flow-matching action head into a stochastic one-step policy by conditioning a single-jump generator on a Brownian path. We derive a conditional action density via our stochastic one-step policy, enabling efficient post-training via likelihood-ratio RL algorithms. We show that distilling VLA-SSD from the $\pi_0$ VLA reduces the action head's forward-pass time by a factor of 7.7, at the cost of a 5.5\% relative decrease in the success rate on the LIBERO benchmark. However, by leveraging the increased sample diversity of VLA-SSD during group generation, post-training a $\pi_{0.5}$ VLA using GRPO yields a 20.6\% relative improvement.

\end{abstract}

\section{Introduction}
\label{sec:introduction}

Vision--language--action (VLA) models have achieved significant advances by leveraging knowledge from pre-trained vision--language models in robotics~\citep{brohan2023rt2,kim2024openvla}. However, state-of-the-art VLAs have yet to achieve comparable generalization to other foundation models due to their reliance on imitation learning from expert demonstrations for adaptation to embodiments and environments~\citep{black2024pi0,intelligence2025pi05,kim2025openvlaoft,nvidia2025groot,shukor2025smolvla}. Frontier flow-based VLAs generate continuous action chunks with a flow-matching action head~\citep{black2024pi0,intelligence2025pi05}. Yet, enabling them to improve via on-policy reinforcement learning (RL) is still limited by the inference time of multi-step ODE integration and a lack of stochasticity.

To improve efficiency, one-step flow-based policies have been introduced for continuous control~\citep{zhan2026mean,ziakas2026fmq}, replacing multi-step ODE integration with a single transport step learned via self-distillation~\citep{song2023consistency,boffi2024flowmaps,geng2025meanflow,boffi2025selfdistillation}. However, in RL methods using multi-sample advantage estimation, such as Group Relative Policy Optimization (GRPO), the baseline is derived directly from a group of trajectories given the current state. Because flow-matching rely entirely on the initial random noise draw for stochasticity, they restrict the action diversity within the sampled group, resulting in a degraded training signal~\citep{shao2024deepseekmath}.  To enable exploration, existing approaches increase sample diversity by introducing an external noise network or reverting to multi-step samplers that inject noise at each integration step~\citep{pirl2025,reinflow2025,flowmapgrpo2026}. 

In this work, we address both the efficiency and stochasticity limitations of current VLAs by learning a stochastic one-step policy via self-distillation.  We propose VLA-SSD, a method that utilizes It\^o flow maps~\citep{pan2026itomaps} to distill a multi-step flow-matching VLA head into a one-step stochastic policy. In contrast to prior work, we introduce stochasticity directly through self-distillation rather than relying on an external noise network or a multi-step sampler. To enable likelihood-ratio policy optimization such as GRPO, we derive a conditional action density that requires only a single additional Jacobian-vector product to approximate. In our experiments, we first distill \nameshort from $\pi_{0.5}$, achieving a $7.7\times$ faster action head with a $0.3$-point drop in success rate, while reducing zero-variance groups from $63\%$ to $35\%$ relative to the original sampler. Leveraging the stochasticity and efficiency of \nameshort, we
post-train it with GRPO, reaching $93.0\%$ on LIBERO, a
$15.9$-point gain over the SFT checkpoint, and matching four-step
GRPO with $\pi_{\mathrm{RL}}$ ($91.5\%$) while generating each action
chunk $3.2\times$ faster during training rollouts.

In summary, our main contributions are as follows:
\begin{itemize}
    \item \textbf{One-Step Stochastic VLA Policies:} We introduce \nameshort, a novel self-distillation framework that converts multi-step flow-matching VLA heads into efficient, one-step stochastic policies using It\^o flow maps.

    \item \textbf{Density Evaluation for Likelihood-Ratio Optimization:} To overcome the lack of tractable likelihoods in flow-based models, we derive a closed-form conditional action density that evaluates efficiently via a single Jacobian-vector product.

    \item \textbf{Efficient Policy Distillation:} We demonstrate that distilling \nameshort\ from $\pi_{0.5}$ yields a $7.7\times$ faster action head with a minimal $0.3$-point performance drop, while mitigating the exploration bottleneck by reducing zero-variance groups from $63\%$ to $35\%$.
    
    \item \textbf{RL Post-Training:} When post-trained with GRPO, \nameshort\ reaches a $93.0\%$ success rate on LIBERO, a $15.9$-point gain over the SFT checkpoint, matching the standard four-step $\pi_{\mathrm{RL}}$ baseline while accelerating training rollouts by $3.2\times$.

\end{itemize}

\begin{figure}[t]
    \centering
    \includegraphics[width=\textwidth]{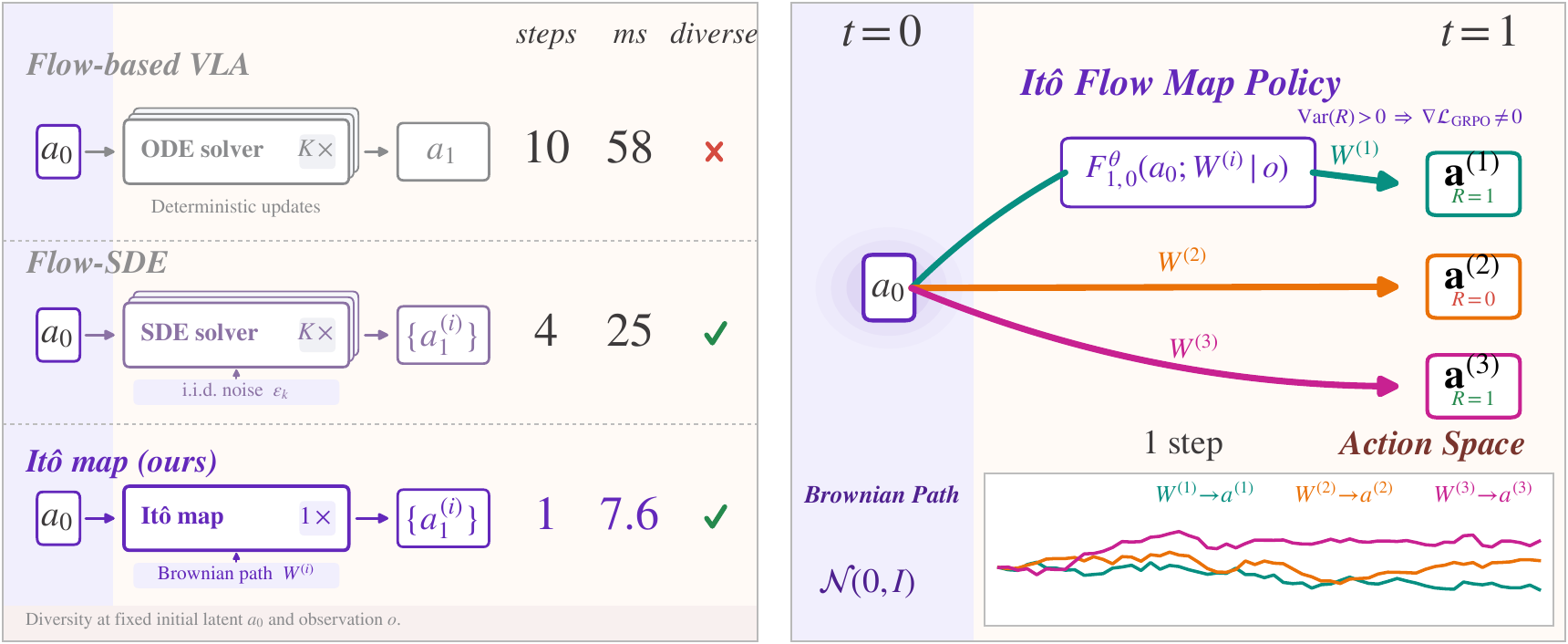}
    \caption{\small \looseness=-1 \textbf{Left:} an ODE sampler and Flow-SDE spend $K$
action-expert evaluations per chunk; \nameshort spends one and keeps the
spread of the SDE. \textbf{Right:} for a fixed $s$ and $a_0$, the
It\^o flow map conditions on a Brownian path $W^{(k)}$, so $G$ paths
give $G$ different one-step actions with different returns, and the
GRPO group carries gradient; at $\sigma{=}0$ all $G$ coincide.}
    \label{fig:overview}
\end{figure}

\section{Preliminaries}
\label{sec:prelim}

\looseness=-1
\xhdr{Markov decision process}
\label{sec:prelim_mdp}
We consider a Markov decision process
(MDP)~\citep{sutton1998reinforcement}
$\gM=(\gS,\gA,r,P,\gamma)$ with state space $\gS$, action space
$\gA\subseteq\R^{d}$, reward function $r(s,a)$, transition kernel
$P(s'\mid s,a)$, and discount factor $\gamma\in[0,1]$.
The state $s\in\gS$ consists of the visual
observations, the proprioceptive state, and the language
instruction; the action $a\in\gA$ is a chunk of $H$
consecutive actions.
A policy $\pi(a\mid s)$ is evaluated by its expected return
$\gJ(\pi)=\E_{\pi}\bigl[\sum_{\tau=0}^{T-1}\gamma^{\tau}
r(s^{\tau},a^{\tau})\bigr]$, where $\tau$ indexes decision steps
and $T$ is the episode horizon.
We consider episodic manipulation tasks with a sparse binary
reward, $r(s^{\tau},a^{\tau})=1$ at the decision step at which the
task is first completed and $0$ otherwise, and take $\gamma=1$, so
that the return of an episode is
$R=\sum_{\tau}r(s^{\tau},a^{\tau})\in\{0,1\}$.
Throughout, $\tau$ denotes MDP time; the flow time $t\in[0,1]$
introduced next indexes the generation of a single action chunk and
is distinct from $\tau$.

\looseness=-1
\xhdr{Flow-matching policies}
\label{sec:prelim_flow} A generative policy $a_1\sim\pi_1(\cdot\mid s)$ can be formulated as a transport from a reference measure $p_0\in\gP(\R^d)$ to a target action
  distribution $p_1\in\gP(\R^d)$. The transport is defined by a state-conditioned flow $\psi_t(\cdot\mid s)$ that solves the following ODE:
\begin{equation}
      \frac{\mathrm d}{\mathrm dt}\psi_t(a_0\mid s)
= v_t\bigl(\psi_t(a_0\mid s)\mid s\bigr),
\qquad \psi_0(a_0\mid s)=a_0,
     \label{eq:flow_policy_ode}
\end{equation}
\looseness=-1
where $v_t$ is a time-dependent velocity field. The terminal map induces the policy $\pi_1(\cdot\mid s)=[\psi_1(\cdot\mid s)]_{\#}p_0$; it is deterministic for fixed $a_0$ and $s$, while sampling $a_0\sim p_0$ induces a distribution over actions. To train the flow policy, we define a conditional probability path $p_t(\cdot\mid z,s)$ using the endpoints $z=(a_0,a_1)$ and the linear interpolation $a_t=(1-t)a_0+ta_1$.
The velocity field is learned by regressing against the target conditional velocity $v_t^*(a_t\mid z,s)$ using the conditional flow-matching objective~\citep{lipman2023flow,albergo2023stochasticinterpolants}:
\begin{equation*}
\gL_{\rm CFM} = \mathbb{E}_{t, q(z\mid s), p_t(a_t | z,s)} \|v_{t}( a_t \mid s) - v^*_t(a_t \mid z,s)\|_2^2 = \mathbb{E}_{t, q(z\mid s), p_t(a_t \mid z,s)} \|v_{t}(a_t \mid s) - (a_1 - a_0)\|_2^2,
\end{equation*}
\looseness=-1
where $q(z\mid s)$ is a coupling over the endpoints
$z=(a_0,a_1)$---e.g., independent coupling
$q(z\mid s)=p_0(a_0)p_1(a_1\mid s)$---and in the last equality
we substitute the analytic conditional velocity
$v_t^*(a_t\mid z,s)=a_1-a_0$ for the linear interpolant.

\xhdr{Flow map policies}
Instead of solving the ODE, we parameterize and learn its two-time
solution operator, referred to as the flow map \citep{song2023consistency,frans2025shortcut,geng2025meanflow,boffi2024flowmaps,albergo2023stochasticinterpolants,ziakas2026fmq}. For
$0\leq r\leq t\leq1$, let
$X_{r,t}(\cdot\mid s):\R^d\to\R^d$ denote the flow map associated
with the ODE in~\cref{eq:flow_policy_ode}. It transports an action
from time $r$ to $t$ and satisfies the jump condition
\begin{equation}
  X_{r,t}(a_r\mid s)=a_t.
  \label{eq:flowmap_jump}
\end{equation}
The one-step policy draws $a_0\sim p_0$ and returns
$a_1=X_{0,1}(a_0\mid s)$, inducing the action distribution
\begin{equation}
  \pi(\cdot\mid s)
  =
  [X_{0,1}(\cdot\mid s)]_{\#}p_0.
  \label{eq:flowmap_policy}
\end{equation}

Following \citet{boffi2024flowmaps}, we parameterize the flow map
using the average action velocity $u_{r,t}$:
\begin{equation}
  X_{r,t}(a_r\mid s)
  =
  a_r+(t-r)u_{r,t}(a_r\mid s),
  \qquad
  u_{r,t}(a_r\mid s)
  =
  \frac{1}{t-r}
  \int_r^t v_\zeta(a_\zeta\mid s)\,\mathrm{d}\zeta.
  \label{eq:flowmap}
\end{equation}
Thus, $X_{r,t}$ represents a finite jump of size $t-r$ along the
ODE trajectory. On the diagonal, the average velocity recovers the
instantaneous velocity,
\begin{equation}
  u_{t,t}(a_t\mid s)=v_t(a_t\mid s).
  \label{eq:flowmap_diagonal}
\end{equation}
This diagonal relation connects the finite-time flow map to the
instantaneous flow-matching velocity. We use it to construct the
training objective for the stochastic map in \cref{sec:distill}.

\xhdr{Group relative policy optimization}
\label{sec:prelim_rl}
GRPO~\citep{shao2024deepseekmath} is an on-policy reinforcement
learning algorithm that replaces the learned value baseline of
actor--critic methods with a Monte Carlo baseline computed within a
group of rollouts.
At each decision step $\tau$ the policy samples an action chunk
$a^{\tau}\sim\pi_\theta(\cdot\mid s^{\tau})$, and the objective of
\cref{sec:prelim_mdp} reduces to $\gJ(\theta)=\E_{\pi_\theta}[R]$
with the binary return $R\in\{0,1\}$.
GRPO samples a group of $N$ trajectories from the same task and
initial state, and assigns to every decision step of trajectory $i$
an advantage $\hat A_i$ obtained by normalizing its return within
the group:

\begin{equation}
    \hat A_i=\frac{R_i-\bar R}{\operatorname{std}(R)+\kappa},
    \qquad
    \bar R=\frac{1}{N}\sum_{j=1}^{N}R_j.
    \label{eq:grpo_adv}
\end{equation}
If all $N$ returns in a group coincide, $\hat A_i=0$ for all members; we term such groups \emph{degenerate}, as they consume rollout computation but contribute no gradient. Using these advantages, GRPO updates the policy by minimizing a PPO-style~\citep{schulman2017ppo} clipped surrogate:
\begin{equation}
    \gL_{\mathrm{clip}}(\theta)
    =-\E\left[\min\left(
        \rho^{\tau}(\theta)\hat A,\;
        \operatorname{clip}\bigl(\rho^{\tau}(\theta),1-\epsilon,1+\epsilon\bigr)
        \hat A\right)\right],
    \quad
    \rho^{\tau}(\theta)
    =\frac{\pi_\theta(a^{\tau}\mid s^{\tau})}
          {\pi_{\theta_{\mathrm{old}}}(a^{\tau}\mid s^{\tau})}.
    \label{eq:clip}
\end{equation}
\looseness=-1

\

\section{Method}
\label{sec:method}

\subsection{Stochastic Flow Map Policies}
\label{sec:distill}

\xhdr{Introducing stochasticity via It\^o maps} While standard flow maps enable one-step generation, their stochasticity is limited to the initial random noise $a_0$. To this
end, we extend the flow map in \cref{eq:flowmap} to an It\^o map, which takes an intermediate action and a realization of a standard Brownian motion $W$ over $[0,1]$ as inputs and predicts a future action on the same stochastic path~\citep{pan2026itomaps}. Evaluated from $r=0$ to $t=1$, it defines a stochastic one-step policy. In practice, we condition the network on a finite-dimensional representation of $W$.

Following the It\^o-map construction of
\citet{pan2026itomaps}, its martingale component $M_t$ captures the Brownian noise accumulated from time 0 to t, weighted by the diffusion profile $\beta(t)$ and scaled by the noise level $\sigma$:
\begin{equation}
  M_t
  =
  \sigma\int_0^t\beta(\zeta)\,\mathrm{d}W_\zeta,
  \qquad
  \Delta M_{r,t}=M_t-M_r,
  \label{eq:method_martingale}
\end{equation}
where $\beta(t)$ is the diffusion profile and $\sigma\geq0$ is the
noise scale. The It\^o map then combines a path-dependent average
drift with the martingale increment $\Delta M_{r,t}$:

\begin{equation}
    F^\theta_{r,t}(a;W\mid s)
    =a+(t-r)\,G^\theta_{r,t}(a;W\mid s)
      +\Delta M_{r,t},
    \label{eq:method_map}
\end{equation}
where $G^\theta_{r,t}$ approximates the path-conditioned average
finite-variation term in the It\^o-map decomposition. At
$\sigma=0$, the martingale $M_t$ vanishes and the stochastic
construction reduces to the deterministic flow map in
\cref{eq:flowmap}.
For each action coordinate, we use the truncated Karhunen--Lo\`eve
expansion $W_t=\sum_{k=1}^{L}\phi_k(t)\xi_k$, where
$\xi\sim\gN(0,I_L)$.
The increment is then linear in the coefficients,
\begin{equation}
    \Delta M_{r,t}=\sigma\,c(r,t)^{\mathsf T}\xi,
    \qquad
    c_k(r,t)=\int_r^t\beta(\zeta)\,\dot\phi_k(\zeta)\,\mathrm{d}\zeta,
    \label{eq:method_increment}
\end{equation}
so $\Delta M_{r,w}+\Delta M_{w,t}=\Delta M_{r,t}$ holds
pathwise for $r\leq w\leq t$, and the same $\xi$ conditions
$G^\theta$ on every subinterval.
In practice, we discretize the time-changed Brownian representation
rather than the driving Brownian motion directly, using
$c_k(r,t)=\phi_k((1-r)^2)-\phi_k((1-t)^2)$
for $\beta(t)=\sqrt{2(1-t)}$; see \cref{app:impl}.
All finite-dimensional density calculations use these coefficients.
For $m=\Delta M_{0,1}$, the stochastic policy is
\begin{equation}
    a_1
    =a_0+G^\theta_{0,1}(a_0;W\mid s)+m.
    \label{eq:method_action}
\end{equation}

\xhdr{Self-distillation}
Starting from the pretrained velocity field, we train the It\^o map
without an external teacher
\citep{boffi2025selfdistillation,ziakas2026fmq}. On the diagonal
$r=t$, the drift $G^\theta_{t,t}$ is independent of the Brownian
path, and we evaluate it using the zero path
$G^\theta_{t,t}(a;0\mid s)$. Compared with standard flow matching,
its regression target includes the score correction of the
marginal-preserving SDE. Under
$\beta(t)=\sqrt{2(1-t)}$, this target becomes
$a_1-(1+\sigma^2)a_0$; the derivation is given in
\cref{app:ito_target}. Following \citet{boffi2024flowmaps}, we consider Lagrangian
(LSD), Eulerian (ESD), and progressive (PSD) self-distillation.
These enforce consistency at the transported endpoint,
at the starting point, and between direct and composed
jumps, respectively (\cref{app:losses}).

In this work, we use Lagrangian self-distillation (LSD)
\citep{boffi2024flowmaps,pan2026itomaps}. At fixed $a_r$ and
Brownian coefficients $\xi$, let
\begin{equation}
  y=F^\theta_{r,t}(a_r;W\mid s)
  \label{eq:ito_transport}
\end{equation}
denote the transported action. LSD matches the endpoint derivative
of the finite-variation component to the diagonal drift evaluated
at $y$. The diagonal loss is
\begin{equation}
  \gL_{\mathrm{Diag}}
  =
  \E\left[
      \left\|
      G^\theta_{t,t}(a_t;0\mid s)
      -\bigl(a_1-(1+\sigma^2)a_0\bigr)
      \right\|_2^2
  \right].
  \label{eq:diag_loss}
\end{equation}
The LSD loss is
\begin{equation}
  \gL_{\mathrm{LSD}}
  =
  \E_{r<t,\,a_r,\,\xi}\left[
      \left\|
      G^\theta_{r,t}(a_r;W\mid s)
      +(t-r)\partial_tG^\theta_{r,t}(a_r;W\mid s)
      -\texttt{sg}\!\left(
          G^\theta_{t,t}(y;0\mid s)
      \right)
      \right\|_2^2
  \right].
  \label{eq:method_losses}
\end{equation}

Here, $\texttt{sg}$ denotes a stop-gradient. The derivative is
taken at fixed $a_r$ and $\xi$ on the finite-variation component
\[
  F^\theta_{r,t}(a_r;W\mid s)-\Delta M_{r,t}
  =
  a_r+(t-r)G^\theta_{r,t}(a_r;W\mid s),
\]
so the martingale increment itself is not differentiated.

At $\sigma=0$, the score correction and martingale component vanish,
and \cref{eq:method_losses} reduces to deterministic flow-map
self-distillation. We use LSD because the Eulerian objective does
not identify the It\^o map~\citep[Prop.~2]{pan2026itomaps}, while
the progressive objective requires two sequential forward passes.
The full objective is
$\gL_{\mathrm{Diag}}+\lambda\gL_{\mathrm{LSD}}$; implementation
details are provided in \cref{app:impl}.

\subsection{On-Policy Optimization of Stochastic Flow Maps}
\label{sec:method_grpo}

\xhdr{Intractable action likelihood} To optimize our policy via likelihood-ratio RL, we need a tractable closed-form evaluation for the conditional action density $\pi_\theta(a_1\mid s)$. Since continuous flow-based models lack this, prior work converts the flow model into a discrete-time Markov process and computes likelihood ratios over $K$ intermediate Gaussian denoising steps rather than the final action~\citep{pirl2025,reinflow2025}. Our approach bypasses this limitation by deriving an auxiliary-variable conditional density. To do so, we isolate the Gaussian component of the Brownian path that generates the martingale increment $m$ in \cref{eq:method_action}, enabling us to treat the one-step action as an invertible transformation of a single Gaussian variable, conditioned on the remaining path randomness.

To isolate the martingale increment $m$ from the rest of the Brownian path, we project the path's random coefficients into two independent components. Let $c=c(0,1)\neq0$ be the full-interval coefficient from \cref{eq:method_increment}, and define its unit direction $e=c/\|c\|_2$. For each action coordinate $j=1,\ldots,d$, the increment is $m_j=\sigma c^{\mathsf T}\xi_j$. We decompose the underlying standard Gaussian $\xi_j$ into a component parallel to $e$ (which perfectly determines $m_j$) and an orthogonal remainder $\xi_{\perp,j}$:
\begin{equation}
  \xi_j
  =
  e\,\frac{m_j}{\sigma\|c\|_2}
  +\xi_{\perp,j},
  \qquad
  \xi_{\perp,j}
  =
  (I_L-ee^{\mathsf T})\xi_j.
  \label{eq:method_latent_split}
\end{equation}
Let $m=(m_1,\ldots,m_d)$ and $\xi_\perp=(\xi_{\perp,1},\ldots,\xi_{\perp,d})$. We group all the randomness that we want to condition on into a single auxiliary variable $h =(a_0,\xi_\perp)$.
Because orthogonal projections of a standard Gaussian are independent, the additive noise follows $m\sim\gN(0,\Sigma I_d)$ with variance $\Sigma =\sigma^2\|c\|_2^2$, completely independent of $\xi_\perp$. Since the initial latent $a_0$ is also drawn independently, $m$ is independent of the entire auxiliary variable $h$. 

Conditioned on $(s,h)$, the Brownian path is fully determined by $m$, so the action in \cref{eq:method_action} is a function of $m$ alone,
\begin{equation}
  a_1=T_\theta(m;s,h)=a_0+g_\theta(m;s,h)+m,
  \label{eq:method_T}
\end{equation}
where $g_\theta(m;s,h)$ is the full-jump drift $G^\theta_{0,1}$ evaluated on the reconstructed path and $J_\theta=\partial g_\theta/\partial m$ is its Jacobian. Given a stored rollout $(s,a_1,h)$, the following proposition gives the density of $a_1$ under updated parameters $\theta$, using the change of variables of invertible residual networks \citep{behrmann2019invertible,chen2019residual}.

\begin{mdframed}[style=MyFrame2]
  \begin{proposition}[Conditional action density]
  \label{thm:conditional_density}
  Assume $\sigma>0$ and $\sup_m\|J_\theta(m;s,h)\|_2<1$. Then $m\sim\gN(0,\Sigma I_d)$ independently of $h$, with $\Sigma=\sigma^2\|c\|_2^2$, the map $T_\theta(\cdot\,;s,h)$ is invertible, and
  \begin{equation}
    \log\pi_\theta(a_1\mid s,h)
    =\log\gN\bigl(m_\theta^\star;0,\Sigma I_d\bigr)
     -\log\det\bigl(I_d+J_\theta(m_\theta^\star;s,h)\bigr),
    \label{eq:method_exact_density}
  \end{equation}
  where $m_\theta^\star=T_\theta^{-1}(a_1;s,h)$.
  \end{proposition}
\end{mdframed}

\begin{proof}[Proof sketch]
The action is generated by $a_1 = a_0 + g_\theta(m;s,h) + m$. The randomness comes from the initial latent $a_0$ and the Brownian coefficients $\xi$. As shown above, splitting $\xi$ isolates the additive noise $m \sim \gN(0, \Sigma I_d)$, which is entirely independent of the auxiliary variable $h = (a_0, \xi_\perp)$. The full proof is in \cref{app:conditional_density_proof}.
\end{proof}

To evaluate \cref{eq:method_exact_density} in practice, we approximate the updated preimage with one fixed-point iteration initialized at the stored rollout increment, and retain the first-order term $\log\det(I_d+J_\theta)\approx\operatorname{tr}(J_\theta)$. We estimate this trace with a single Hutchinson probe, requiring one additional Jacobian-vector product, and reuse the same probe for the old and updated policies to reduce variance in their score difference. The resulting score approximates the conditional log-density; the inverse update and scoring formula are provided in \cref{sec:scoring,eq:method_approx_density}.

\xhdr{Likelihood ratios with auxiliary variables}
\label{sec:aux}
\Cref{thm:conditional_density} gives the action density conditioned on the auxiliary variable $h=(a_0,\xi_\perp)$, rather than the marginal density $\pi_\theta(a_1\mid s)$. Let $\nu(h)$ denote the distribution of $h$, which is independent of $\theta$. The marginal policy is
\begin{equation}
  \pi_\theta(a_1\mid s)
  =
  \int
  \nu(h)\pi_\theta(a_1\mid s,h)\,\mathrm{d}h.
  \label{eq:method_marginal_policy}
\end{equation}
We equivalently define the augmented policy
\begin{equation}
  \bar\pi_\theta(h,a_1\mid s)
  =
  \nu(h)\pi_\theta(a_1\mid s,h).
  \label{eq:method_augmented_policy}
\end{equation}
The augmented policy samples the pair $(h,a_1)$, while the environment receives only $a_1$. Marginalizing out $h$ therefore recovers \cref{eq:method_marginal_policy}, so $\bar\pi_\theta$ and $\pi_\theta$ induce the same trajectory distribution and expected return $\gJ$.

Because $\nu(h)$ is independent of $\theta$,
\begin{equation}
  \nabla_\theta\log\bar\pi_\theta(h,a_1\mid s)
  =\nabla_\theta\log\pi_\theta(a_1\mid s,h),
  \qquad
  \frac{\bar\pi_\theta(h,a_1\mid s)}{\bar\pi_{\theta_{\mathrm{old}}}(h,a_1\mid s)}
  =\frac{\pi_\theta(a_1\mid s,h)}{\pi_{\theta_{\mathrm{old}}}(a_1\mid s,h)}.
  \label{eq:method_augmented_ratio}
\end{equation}
Consequently, before the approximations in \cref{sec:scoring}, the policy-gradient estimator constructed from the conditional density is unbiased for $\nabla_\theta\gJ$. The augmented and marginal formulations optimize the same expected-return objective, although their likelihood-ratio variances and clipped surrogate objectives can differ.

The auxiliary variable sampled during rollout is stored and reused when rescoring the recorded action; the Brownian path is never resampled. Using the approximate score from \cref{eq:method_approx_density}, the likelihood ratio is
\begin{equation}
  \widetilde\rho_i^\tau(\theta)
  =
  \exp\!\left[
      \widetilde\ell_\theta
      (a_i^\tau\mid s_i^\tau,h_i^\tau)
      -
      \widetilde\ell_{\theta_{\mathrm{old}}}
      (a_i^\tau\mid s_i^\tau,h_i^\tau)
  \right],
  \label{eq:method_approx_ratio}
\end{equation}
where $i$ indexes trajectories and $\tau$ indexes decision steps.

\xhdr{GRPO}
For each update, we collect groups of $N$ trajectories from the same task and initial environment state. Group members use independent initial latents $a_0$ and Brownian coefficients $\xi$, and the auxiliary variables sampled at every decision step are stored for policy rescoring. Episode returns are converted into the group-normalized advantages in \cref{eq:grpo_adv}. The Brownian path increases action diversity beyond that provided by $a_0$ alone. Without this additional stochasticity, the policy more frequently produces zero-variance groups, for which all advantages vanish; \cref{sec:exp_exploration} measures this effect. We minimize the clipped GRPO surrogate in \cref{eq:clip} using the approximate likelihood ratio $\widetilde\rho_i^\tau(\theta)$ from \cref{eq:method_approx_ratio}. The stored auxiliary variable $h_i^\tau$ is reused when evaluating both the old and updated policies. Training details are provided in \cref{app:impl}.

\section{Experiments}
\label{sec:experiments}

The experiments evaluate the end-to-end performance of \nameshort and analyze its underlying mechanics. We first evaluate the final on-policy reinforcement learning success rates and computational efficiency against multi-step baselines (\cref{sec:exp_rl}). We then analyze the source of the one-step policy's sample diversity and its impact on the GRPO signal (\cref{sec:exp_exploration}). Finally, we ablate the initial self-distillation phase to isolate the impact of different consistency objectives on the base policy (\cref{sec:exp_ablation}).

\subsection{Experimental Setup}
\label{sec:exp_setup}

\xhdr{Model and checkpoints}
We use $\pi_{0.5}$~\citep{intelligence2025pi05}, a
3.3B-parameter VLA model with a 300M-parameter action expert
serving as the flow-matching head described in
\cref{sec:method}. The vision--language backbone is kept
frozen throughout training.
All runs start from the SFT checkpoints released with
$\pi_{\mathrm{RL}}$~\citep{pirl2025}, so differences from its
published numbers come from the post-SFT procedure alone.
On LIBERO, reinforcement learning starts from the few-shot
checkpoint (77.1\%), as in $\pi_{\mathrm{RL}}$; on MetaWorld the
released checkpoint scores 51.8\% under our protocol.

\xhdr{Benchmarks and protocol}
LIBERO~\citep{libero2023} is evaluated on its four ten-task suites
(spatial, object, goal, long) with 500 episodes per suite; a
four-suite mean therefore averages 2{,}000 episodes and has a
standard error of 0.5--0.7 points.
MetaWorld MT50~\citep{metaworld2020} is evaluated on 50 tasks
$\times$ 10 fixed reset states at horizon 100, following
$\pi_{\mathrm{RL}}$.
Unless stated otherwise, success rates are deterministic evaluations.

\xhdr{Baselines}
We compare against $\pi_{\mathrm{RL}}$'s GRPO with Flow-SDE and PPO
with Flow-Noise at their operating depth $K{=}4$, using the results that work
reports on the same checkpoints and protocol; against the
deterministic flow map distilled with the same recipe at
$\sigma=0$; and against the released ODE sampler at $K{=}4$ and $K{=}10$.
We also evaluate a four-step ReinFlow-style PPO baseline (Appendix~\ref{app:reinflow}).%

\xhdr{Training}
Distillation uses the Lagrangian objective, with the sampling
interval gradually expanded over the first quarter of training,
$L{=}5$ Karhunen--Lo\`eve modes, and $\sigma{=}0.1$.
We use the profile $\beta(t)=\sqrt{2(1-t)}$, for which the
diagonal correction in \cref{eq:diag_loss} equals
the constant $\sigma^2$.
GRPO runs in RLinf~\citep{rlinf2025} on four A100 GPUs with
groups of $N{=}8$ rollouts sharing a task and initial state.
\Cref{app:impl} provides all settings.

\subsection{Main Results: On-Policy RL Performance}
\label{sec:exp_rl}

On both benchmarks in \cref{tab:main}, the performance improvement from GRPO far exceeds the initial drop caused by one-step distillation. The final policy improves over the released SFT checkpoint by 15.9 percentage points on LIBERO and 20.2 points on MetaWorld. Under GRPO, our one-step policy achieves 93.0\% on LIBERO, outperforming $\pi_{\mathrm{RL}}$'s four-step GRPO policy ($91.5\%$). While four-step PPO achieves a higher 98.3\% on LIBERO, it requires the substantial overhead of training an additional vision--language critic. On MetaWorld, our one-step GRPO policy achieves 72.0\%, outperforming the four-step PPO baseline ($70.7\%$).

\begin{table}[t]
\centering
\caption{Main results. Success on the LIBERO four-suite mean and
MetaWorld MT50 (ours at $\sigma{=}0$); passes is action-expert
evaluations per chunk, forward its time per chunk on one A100.
$\pi_{\mathrm{RL}}$ rows are its published numbers from the same
SFT checkpoints. Bold is best per column.}
\label{tab:main}
\small
\begin{tabular}{llcccc}
\toprule
Policy & Sampler & Passes & Forward (ms) & LIBERO & MetaWorld \\
\midrule
$\pi_{0.5}$ SFT & ODE, $K{=}10$ & 10 & 58.4 & 77.1 & 51.8 \\
$\pi_{\mathrm{RL}}$, GRPO & SDE, $K{=}4$ & 4 & 24.5 & 91.5 & --- \\
$\pi_{\mathrm{RL}}$, PPO & SDE, $K{=}4$ & 4 & 24.5 & \textbf{98.3} & 70.7 \\
\midrule
It\^o student (ours) & It\^o, $K{=}1$ & 1 & 7.6 & 68.6 & 49.0 \\
It\^o + GRPO (ours) & It\^o, $K{=}1$ & 1 & 7.6 & 93.0 & \textbf{72.0} \\
\bottomrule
\end{tabular}
\end{table}

\subsection{Analysis: Stochasticity and Sample Diversity}
\label{sec:exp_exploration}

\xhdr{Source of sample diversity}
\Cref{tab:diversity} distinguishes variation induced by the
initial latent from variation introduced during generation
by fixing $a_0$ (\cref{fig:diversity}).
Both the ODE sampler and the deterministic map are
deterministic conditional on $a_0$, and their effective rank
drops to zero when $a_0$ is fixed.
With independently sampled latents, the ODE sampler has an
effective rank of 3.2.
The It\^o map retains a similar rank with fixed and free
latents (18.5 and 18.2), showing that substantial diversity
can be generated through the Brownian path alone.
Its fixed-latent rank exceeds that of four-step Flow-SDE
(14.3) and matches that of isotropic Gaussian output noise.
Their effects on return variation nevertheless differ,
as examined next.

\xhdr{Zero-variance groups}
The group-relative reward term in GRPO provides no learning
signal when all members of a group obtain identical returns.
Success rates are similar across samplers (83.1--84.6\%),
whereas the fraction of zero-variance groups varies.
The ODE sampler and deterministic map produce zero return
variance in 62.5\% and 56.2\% of groups, respectively.
Isotropic output noise reduces this fraction to 47.9\%,
and four-step Flow-SDE to 45.8\%.
The It\^o map yields the lowest observed fraction, 35.4\%,
despite having a similar effective rank to the isotropic
control.
This suggests that its sample diversity translates more
effectively into within-group return variation.
Under binary rewards, a lower zero-variance fraction
corresponds to a larger proportion of groups providing
a group-relative learning signal

\xhdr{Retention under training noise}
Exploration noise is useful only if the noisy policy still
succeeds often enough for groups to contain both outcomes.
On MetaWorld the It\^o student attains 36.8\% under the training
noise $\sigma{=}0.1$ against 49.0\% deterministically, a retention
of 75\%; after GRPO the figures are 55.4\% and 72.0\%, a retention
of 77\%.
For comparison, $\pi_{\mathrm{RL}}$ reports that Flow-SDE at
$K{=}1$ retains 14.7\% of its deterministic success and 86\% at
$K{=}4$, which is why that method is trained at four steps.

\begin{minipage}[t]{0.5\textwidth}
    \vspace{0pt}
    \centering
    \begingroup
    \scriptsize
    \setlength{\tabcolsep}{3pt}
    \renewcommand{\arraystretch}{1.65}
    \resizebox{\linewidth}{!}{%
        \begin{tabular}{@{}lcccc@{}}
        \toprule
        & \multicolumn{2}{c}{Effective rank}
        & \multicolumn{2}{c}{GRPO groups} \\
        \cmidrule(lr){2-3}\cmidrule(lr){4-5}
        Sampler & Free $a_0$ & Fixed $a_0$
        & SR (\%) & Zero-var. \\
        \midrule
        $\pi_{0.5}$ ODE, $K{=}10$
        & 3.2 & 0.0 & 84.4 & 0.625 \\
        Flow-SDE, $K{=}4$
        & 13.7 & 14.3 & 84.6 & 0.458 \\
        \midrule
        Deterministic map, $K{=}1$
        & 10.3 & 0.0 & 84.4 & 0.562 \\
        \quad + $\gN(0,0.1^2 I)$ noise
        & 18.5 & 18.5 & 84.4 & 0.479 \\
        It\^o map, $K{=}1$, $\sigma{=}0.1$
        & 18.2 & 18.5 & 83.1 & \textbf{0.354} \\
        \bottomrule
        \end{tabular}%
    }
    \endgroup
\end{minipage}\hfill%
\begin{minipage}[t]{0.5\textwidth}
    \vspace{0pt}
    \centering
    \sbox{\diversitybox}{%
        \includegraphics{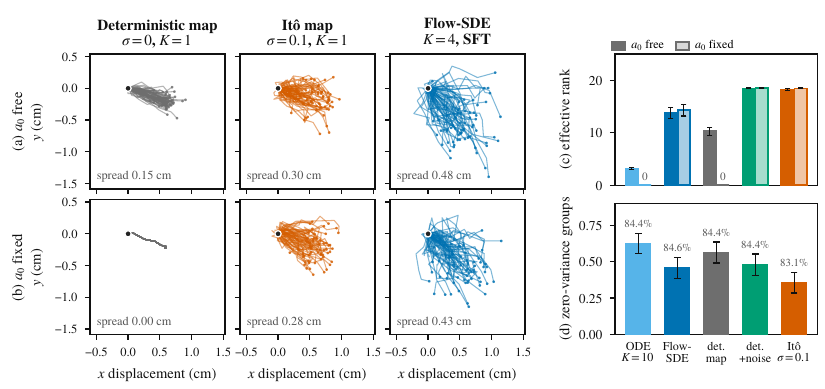}%
    }
    \setlength{\diversitytrim}{0.35\wd\diversitybox}
    \edef\diversityimage{%
        \noexpand\includegraphics[
            width=\the\linewidth,
            trim={0 5bp \the\diversitytrim\space 8.5bp},
            clip
        ]{figures/diversity.pdf}%
    }
    \diversityimage
\end{minipage}

\par\medskip

\begin{minipage}[t]{0.54\textwidth}
    \vspace{0pt}
    \setlength{\abovecaptionskip}{0pt}
    \captionof{table}{Sample diversity and GRPO signal.
Effective rank measures action diversity with free or fixed
initial latent $a_0$; zero-variance is the fraction of rollout
groups with identical returns. SR denotes success rate.
Multi-step samplers use the SFT checkpoint; one-step samplers
use its distilled student.}
\label{tab:diversity}
\label{tab:groups}
\end{minipage}\hfill%
\begin{minipage}[t]{0.44\textwidth}
    \vspace{0pt}
    \setlength{\abovecaptionskip}{0pt}
    \captionof{figure}{End-effector trajectories with
(a) independent and (b) shared initial latents $a_0$.
Fixing $a_0$ collapses the deterministic samples, while
the stochastic samplers retain diversity.
The distilled models shown use an earlier student checkpoint.}
\label{fig:diversity}
\end{minipage}

\textbf{Exploration during GRPO.}
All runs on each benchmark started from the same distilled checkpoint. We evaluated success after 100 optimizer updates with exploration noise turned off. First, we varied only the rollout noise scale. The default $\sigma=0.10$ gave the highest success rate on both benchmarks among the three scales tested. Second, we trained a one-step control with the same GRPO settings, disabling the Brownian-path input and increment and adding independent isotropic Gaussian noise to the output. The It\^o policy led by 2.0 percentage points on LIBERO and 0.4 on MT50.
\par\smallskip
{\centering\small
\captionof{table}{Effect of rollout noise scale and noise source on task success after 100 GRPO optimizer updates.}
\label{tab:grpo-exploration}
\begin{tabular}{@{}lrr@{}}
\toprule
Training condition & LIBERO success (\%) & MT50 success (\%) \\
\midrule
\multicolumn{3}{l}{\emph{(a) Rollout noise scale for the It\^o policy}} \\
$\sigma=0.05$ & 69.10 & 58.80 \\
$\sigma=0.10$ (default) & 73.50 & 59.40 \\
$\sigma=0.20$ & 65.15 & 55.20 \\
\midrule
\multicolumn{3}{l}{\emph{(b) One-step GRPO policies}} \\
It\^o map & 73.50 & 59.40 \\
Isotropic output noise & 71.50 & 59.00 \\
\bottomrule
\end{tabular}\par}

\subsection{Ablation: Impact of Self-Distillation}
\label{sec:exp_distill}
\label{sec:exp_ablation}

\xhdr{Distillation loss}
Distilled from the full-demonstration $\pi_{0.5}$ checkpoint, the
one-step It\^o student attains 96.55\% on the LIBERO four-suite
mean against the ten-step teacher's 96.85\%, with the residual
concentrated on the long-horizon suite.
Evaluated at $K{=}1$ and $K{=}5$, the same checkpoint varies by
less than $\pm2$ points, indicating that the student approximates
the solution operator over the interval rather than a single
endpoint.
The loss is larger when the teacher is weaker: 8.5 points from the
few-shot LIBERO checkpoint ($77.1\to68.6$) and 2.8 points on
MetaWorld ($51.8\to49.0$), the two starting points of
\cref{sec:exp_rl}.

The LIBERO comparison in \cref{tab:main} uses ten ODE steps for the SFT teacher (77.1\%) and one step for the distilled student (68.6\%). Holding sampling depth fixed at one step, self-distillation raises pre-RL success from 40.55\% to 68.60\% (\cref{app:directgrpo}).

The order of training also matters: distilling from the converged
SFT checkpoint rather than jointly from the $\pi_{0.5}$ base
raised a matched-budget student from 12.2\% to 42.7\% at $K{=}1$.

\begin{table}[!ht]
\centering
\caption{One-step distillation from $\pi_0$ on LIBERO, 500 episodes
per suite, all students evaluated at $K{=}1$. The teacher is
$\pi_{\mathrm{RL}}$'s full-dataset SFT $\pi_0$ checkpoint at
$K{=}10$; the deterministic map under ESD is
MeanFlow~\citep{geng2025meanflow}. Forward is the action head's time
per chunk on one A100. Bold marks the best objective for each map;
the standard error on each mean is about 0.7 points.}
\label{tab:distill}
\small
\begin{tabular}{llcccccc}
\toprule
Map & Objective & Forward (ms) & spatial & object & goal & long & mean \\
\midrule
$\pi_0$ teacher, $K{=}10$ & --- & 58.4 & 96.8 & 98.8 & 95.8 & 85.2 & 94.15 \\
\midrule
Deterministic flow map & LSD & 7.6 & 96.2 & \textbf{96.0} & \textbf{94.0} & \textbf{79.0} & \textbf{91.30} \\
 & PSD & 7.6 & \textbf{96.4} & 95.4 & 93.6 & 77.8 & 90.80 \\
 & ESD & 7.6 & 90.2 & 76.8 & 77.0 & 26.0 & 67.50 \\
\midrule
It\^o flow map, $\sigma{=}0.1$ & LSD & 7.6 & 94.2 & \textbf{93.6} & \textbf{91.0} & \textbf{77.0} & \textbf{88.95} \\
 & PSD & 7.6 & \textbf{96.6} & 91.2 & 90.8 & 73.2 & 87.95 \\
 & ESD & 7.6 & 85.4 & 76.8 & 62.6 & 37.2 & 65.50 \\
\bottomrule
\end{tabular}
\end{table}

\xhdr{Consistency objective and map}
\Cref{tab:distill} compares deterministic and It\^o maps
under three consistency objectives using the $\pi_0$ teacher.
The Lagrangian and progressive objectives yield one-step
performance reductions of 2.9--6.2 percentage points,
with the largest reductions on the long-horizon suite.
The Eulerian objective exhibits substantially larger
reductions, consistent with reduced accuracy of first-order
extrapolation from the diagonal as $|t-r|$ increases.
LSD achieves the highest mean success for both maps
and is used throughout.
Under LSD, the It\^o map achieves 88.95\%, compared with
91.30\% for the deterministic map, indicating a
2.35-percentage-point performance trade-off associated
with the stochastic formulation.
Its exploration properties are analyzed
in \cref{sec:exp_exploration}.

\section{Related Work}
\label{sec:related_work}

\subsection{Reinforcement fine-tuning of generative policies}
Building on pretrained
VLAs~\citep{brohan2023rt2,kim2024openvla,black2024pi0,intelligence2025pi05},
online RL has been explored for autoregressive action
generation~\citep{vlarl2025,simplevlarl2025}.
For diffusion and flow policies, DPPO optimizes the denoising
chain with PPO~\citep{ren2025dppo}, while ReinFlow and
$\pi_{\mathrm{RL}}$ introduce stochastic denoising transitions
for likelihood-based fine-tuning, retaining multi-step
generation~\citep{reinflow2025,pirl2025}.
In image generation, Flow-GRPO and DanceGRPO apply GRPO through
ODE-to-SDE conversion~\citep{liu2025flowgrpo,xue2025dancegrpo};
Flow-Map GRPO extends this approach to few-step generators
through anchored stochastic composition~\citep{flowmapgrpo2026}.
Complementary critic-based approaches optimize a fixed policy's
latent-noise distribution (DSRL)~\citep{dsrl2025}, update flow
maps through critic-guided trust regions (Flow Map
$Q$-Guidance)~\citep{ziakas2026fmq}, or combine value calibration
with action filtering (FORCE)~\citep{force2026}.
Our work enables GRPO fine-tuning of a one-step stochastic VLA
through its conditional action density, avoiding a multi-step
denoising chain.

\subsection{One-step generation and stochastic flow maps}
Diffusion and flow policies model multimodal robot
actions through iterative
sampling~\citep{chi2023diffusionpolicy,black2024pi0}.
Consistency models, shortcut models, MeanFlow, and flow-map
methods replace iterative integration with learned finite-time
transport~\citep{song2023consistency,frans2025shortcut,geng2025meanflow,boffi2024flowmaps,albergo2023stochasticinterpolants},
including self-distillation from the map's own
diagonal~\citep{boffi2025selfdistillation}.
In robotics, Consistency Policy and SnapFlow distill policies
into few-step or one-step
samplers~\citep{prasad2024consistencypolicy,snapflow2026},
while Let It Be Simple enables one-step decoding through
high-noise training~\citep{letitbesimple2026}.
These maps are deterministic given the initial noise.
Stochastic extensions include Meta Flow Maps for
reward-conditioned posterior sampling~\citep{potaptchik2026mfm}
and It\^o maps for Brownian-path-conditioned SDE
transitions~\citep{pan2026itomaps}.
We build on It\^o maps to combine one-step VLA generation
with explicit exploration and conditional-density-based
on-policy optimization.

\section{Conclusion}
\label{sec:conclusion}

We presented \nameshort, which distills a flow-matching VLA into a one-step stochastic policy based on It\^o flow maps and fine-tunes it with GRPO. The Brownian path provides exploration with an explicit noise scale, and the derived conditional density makes the policy usable in likelihood-based on-policy updates. On LIBERO and MetaWorld MT50, the result matches multi-step GRPO fine-tuning with $\pi_{\mathrm{RL}}$ using just one action-expert evaluation per chunk instead of four. Regarding limitations, our approximate likelihood scoring relies on a first-order truncation of the log-determinant. In addition, while GRPO avoids the heavy memory overhead of a vision--language critic, it does not match the peak absolute sample efficiency of critic-based PPO on LIBERO. However, because physical robot rollouts operate in real-time and constitute the primary bottleneck of on-policy RL, the $3.2\times$ faster rollout generation provided by \nameshort is particularly valuable for real-world deployment. Translating this framework to real-robot fine-tuning is the natural next step. Finally, investigating adaptive step sizing---where the model defaults to the one-step It\^o map but allocates additional sampling steps for complex, high-precision maneuvers---presents an exciting direction for bridging one-step and multi-step continuous control.

\ificlrfinal
\section*{Acknowledgments}
This work was supported by UKRI (EP/Y037111/1) as part of the ProSafe project (EU Horizon 2020, MSCA, grant no.\ 101119358).
\fi

\bibliography{iclr2027_conference}
\bibliographystyle{iclr2027_conference}

\appendix

\section{Consistency Objectives for Flow-Map Distillation}
\label{app:losses}
 
This appendix states the three self-distillation objectives of
\cref{sec:distill} in full, in the notation of \cref{sec:prelim},
and records the properties that motivate the choice made in the
main text.
Throughout, the flow map is parameterized as
$X^\theta_{r,t}(a_r\mid s)=a_r+(t-r)\,u^\theta_{r,t}(a_r\mid s)$
for $0\leq r\leq t\leq1$, the diagonal $u^\theta_{t,t}$ is trained
by $\gL_{\mathrm{Diag}}$ in \cref{eq:diag_loss}, and $\texttt{sg}$
denotes a stop-gradient; conditioning on $s$ and the superscript
$\theta$ are suppressed where unambiguous.
 
\xhdr{Origin of the conditions.}
The exact solution operator $X_{r,t}$ of the ODE
$\frac{\mathrm{d}}{\mathrm{d}t}a_t=v_t(a_t)$ satisfies the jump
condition $X_{r,t}(a_r)=a_t$ and the semigroup property
$X_{w,t}\circ X_{r,w}=X_{r,t}$ for $r\leq w\leq t$.
Differentiating the jump condition with respect to the endpoint
$t$ gives $\partial_t X_{r,t}(a_r)=v_t(X_{r,t}(a_r))$; differentiating
with respect to the start time $r$ and using
$\frac{\mathrm{d}}{\mathrm{d}r}X_{r,t}(a_r)=0$ gives
$\partial_r X_{r,t}(a_r)+\nabla X_{r,t}(a_r)\,v_r(a_r)=0$.
Each of these identities, together with the semigroup property,
yields a residual that vanishes for the exact map and can be
minimized for the parameterized map, with the diagonal
$u_{t,t}=v_t$ serving as the
teacher~\citep{boffi2024flowmaps,boffi2025selfdistillation,ziakas2026fmq}.
 
\xhdr{Lagrangian self-distillation (LSD)}
Writing $y=X_{r,t}(a_r)$ for the transported state,
\begin{equation}
    \gL_{\mathrm{LSD}}
    =\E_{r<t,\,a_r}\left[
      \left\|\partial_t X_{r,t}(a_r)
       -\texttt{sg}\bigl(u_{t,t}(y)\bigr)\right\|_2^2\right],
    \qquad
    \partial_t X_{r,t}=u_{r,t}+(t-r)\,\partial_t u_{r,t}.
    \label{eq:app_lsd}
\end{equation}
The condition matches the endpoint derivative of the map to the
instantaneous velocity evaluated at the transported state, so the
map is supervised along the trajectory it generates.
The time derivative $\partial_t u_{r,t}$ is obtained by a
Jacobian-vector product with respect to the scalar $t$, and the
target requires one additional evaluation of the diagonal at $y$.
 
\xhdr{Eulerian self-distillation (ESD)}
\begin{equation}
    \gL_{\mathrm{ESD}}
    =\E_{r<t,\,a_r}\left[
      \left\|\partial_r X_{r,t}(a_r)
       +\texttt{sg}\bigl(\nabla X_{r,t}(a_r)\,u_{r,r}(a_r)\bigr)\right\|_2^2\right],
    \qquad
    \partial_r X_{r,t}=-u_{r,t}+(t-r)\,\partial_r u_{r,t}.
    \label{eq:app_esd}
\end{equation}
The condition transports the map along the characteristics of the
diagonal field at the start point, so both the residual and its
target are evaluated at $a_r$ rather than at the transported state.
The spatial term $\nabla X_{r,t}(a_r)\,u_{r,r}(a_r)$ is a
Jacobian-vector product in the action argument.
For the average-velocity parameterization, removing the
stop-gradient and expanding
$\nabla X_{r,t}\,u_{r,r}=u_{r,r}+(t-r)\,\nabla u_{r,t}\,u_{r,r}$
shows that the ESD gradient coincides with that of
MeanFlow~\citep{geng2025meanflow} up to the choice of teacher for
the instantaneous velocity; see \citet[App.~B]{ziakas2026fmq}.
 
\xhdr{Progressive self-distillation (PSD)}
\begin{equation}
    \gL_{\mathrm{PSD}}
    =\E_{r<t,\,a_r,\,\alpha}\left[
      \left\|X_{w,t}\bigl(X_{r,w}(a_r)\bigr)
       -\texttt{sg}\bigl(X_{r,t}(a_r)\bigr)\right\|_2^2\right],
    \qquad
    w=(1-\alpha)\,r+\alpha\,t,\;\;\alpha\sim\gU[0,1].
    \label{eq:app_psd}
\end{equation}
The condition enforces the semigroup property directly: a jump
from $r$ to $t$ must agree with the composition of two shorter
jumps through an intermediate time $w$.
It requires no derivative of the network but two sequential
forward passes on the left-hand side, and it constrains the
off-diagonal map only through other off-diagonal evaluations, so
the diagonal enters indirectly.

\section{Diagonal Target of the It\^o Map}
\label{app:ito_target}

Throughout this derivation, we fix the conditioning state $s$
and assume independent endpoint coupling,
$q(a_0,a_1\mid s)=p_0(a_0)p_1(a_1\mid s)$,
with $p_0=\mathcal N(0,I_d)$.
We suppress the dependence on $s$ below.

The diagonal loss in \cref{eq:diag_loss} regresses the
zero-path diagonal drift onto the drift of the SDE that shares
marginals with the flow-matching ODE.
For the interpolant $a_t=(1-t)\,a_0+t\,a_1$, the marginal path
$p_t$ of the ODE $\frac{\mathrm{d}}{\mathrm{d}t}a_t=v_t(a_t)$ is
also the marginal path of the SDE
\begin{equation}
    \mathrm{d}a_t
    =\Bigl[v_t(a_t)+\tfrac12\,\sigma^2\beta(t)^2\,\nabla\log p_t(a_t)\Bigr]\mathrm{d}t
     +\sigma\beta(t)\,\mathrm{d}W_t,
    \label{eq:app_sde}
\end{equation}
for any diffusion profile $\sigma\beta(t)$, since the added
diffusion and the score correction cancel in the Fokker--Planck
equation.
Conditioned on the endpoint $a_1$, the interpolant is Gaussian
with mean $t\,a_1$ and covariance $(1-t)^2 I_d$, so the conditional
score is $\nabla\log p_t(a_t\mid a_1)=-(a_t-t\,a_1)/(1-t)^2=-a_0/(1-t)$.
Substituting the conditional velocity $a_1-a_0$ and the conditional
score gives the regression target
\begin{equation}
    a_1-a_0-\frac{\sigma^2\beta(t)^2}{2(1-t)}\,a_0
    =a_1-\Bigl(1+\frac{\sigma^2\beta(t)^2}{2(1-t)}\Bigr)a_0,
    \label{eq:app_target}
\end{equation}
which is the target in \cref{eq:diag_loss}; as for
conditional flow matching, regressing onto the conditional target
recovers the marginal drift in expectation.
The coefficient remains bounded as $t\to1$ provided that
$\beta(t)^2/(1-t)$ remains bounded.
With the profile $\beta(t)=\sqrt{2(1-t)}$ used in our experiments,
the coefficient reduces to the constant $1+\sigma^2$, and the
Brownian variance accumulated over $[0,1]$ is
$\sigma^2\int_0^1\beta(t)^2\,\mathrm{d}t=\sigma^2$ in the continuum;
\cref{app:impl} gives the truncated value used in practice.

\xhdr{Density diagnostics}
At the end of distillation the Brownian pathway's output projection
is of order $3\times10^{-4}$; consequently the variance
$2\|(J_\theta+J_\theta^{\mathsf T})/2\|_F^2
\leq 2\|J_\theta\|_F^2$ of the single-probe trace estimator and the
truncation error
$\leq \|J_\theta\|_F^2/[2(1-\|J_\theta\|_2)]$
of the first-order log-determinant in
\cref{eq:method_approx_density}, under $\|J_\theta\|_2<1$, are
both below $10^{-6}$ nats per chunk.
At the first minibatch of every run the ratio
$\widetilde\rho_{i}^{\tau}$ equals one to floating-point precision and
the mean Jacobian correction is indistinguishable from zero.
These values are measured at the start of training; the pathway
norm was not logged during reinforcement learning.

\section{Proof of the Conditional-Density Proposition}
\label{app:conditional_density_proof}

\begin{proof}[Proof of \cref{thm:conditional_density}]
For each action coordinate,
$e^{\mathsf T}\xi_j$ and
$(I_L-ee^{\mathsf T})\xi_j$ are independent Gaussian projections.
Since
$m_j=\sigma\|c\|_2e^{\mathsf T}\xi_j$, the coordinates of $m$ are
independent and
$m\sim\gN(0,\Sigma I_d)$, where
$\Sigma=\sigma^2\|c\|_2^2$. Because $a_0$ is sampled independently
of the Brownian coefficients, $m$ is independent of
$h=(a_0,\xi_\perp)$.

Fix $(s,h)$ and an action $a_1$. A preimage of $a_1$ under $T_\theta$
satisfies
\[
  m=a_1-a_0-g_\theta(m;s,h).
\]
The right-hand side is a contraction under the assumption of the
theorem, so the Banach fixed-point theorem guarantees a unique
solution $m_\theta^\star$. Therefore,
$T_\theta(\cdot;s,h)$ is invertible. Its Jacobian is
\[
  \frac{\partial T_\theta}{\partial m}
  =
  I_d+J_\theta(m;s,h),
\]
which is nonsingular because
$\|J_\theta(m;s,h)\|_2<1$. Applying the change-of-variables formula
to $m\sim\gN(0,\Sigma I_d)$ gives
\cref{eq:method_exact_density}.
\end{proof}

The proposition provides the exact density conditioned on the auxiliary
variable $h$; the marginal action density $\pi_\theta(a_1\mid s)$
remains intractable. We next approximate the inverse and
log-determinant required to evaluate the conditional density.

\xhdr{Approximating policy scoring}
\label{sec:scoring}
After a policy update, the stored increment $m^{(0)}$ need not
generate the recorded action under the updated parameters. The
updated preimage satisfies the fixed-point equation
\begin{equation}
  m
  =
  a_1-a_0-g_\theta(m;s,h).
  \label{eq:method_inverse_fixed_point}
\end{equation}
Starting from the stored increment $m^{(0)}$, we approximate the
updated preimage using one fixed-point iteration:
\begin{equation}
  m^{(1)}_\theta
  =
  a_1-a_0-g_\theta(m^{(0)};s,h).
  \label{eq:method_inverse_approx}
\end{equation}

Under the contraction assumption of
\cref{thm:conditional_density}, the Jacobian correction to
the conditional log-density admits the convergent trace
expansion
\begin{equation}
  \mathcal C(J)
  =
  \sum_{k=1}^{\infty}
  \frac{(-1)^k}{k}\operatorname{tr}(J^k),
  \qquad
  J=J_\theta(m_\theta^\star;s,h).
  \label{eq:method_logdet_series}
\end{equation}
We retain its first-order term and estimate the trace using a
Hutchinson probe $\omega\sim\gN(0,I_d)$:
\begin{equation}
  \begin{aligned}
  \widetilde{\ell}_\theta(a_1\mid s,h)
  ={}&
  -\frac{\|m^{(1)}_\theta\|_2^2}{2\Sigma}
  -\frac{d}{2}\log(2\pi\Sigma)\\
  &-
  \omega^{\mathsf T}
  J_\theta(m^{(0)};s,h)\omega.
  \end{aligned}
  \label{eq:method_approx_density}
\end{equation}
The final term requires one Jacobian-vector product. At
$\theta=\theta_{\mathrm{old}}$, the fixed-point update recovers the
stored increment exactly,
$m^{(1)}_{\theta_{\mathrm{old}}}=m^{(0)}$. We use the same probe
$\omega$ for the old and updated policies to reduce variance in the
likelihood ratio. Thus,
\cref{eq:method_approx_density} is used as an approximate policy
score rather than an exact likelihood.

\begin{minipage}[t]{0.62\linewidth}
\vspace{0pt}\raggedright
\textbf{Scoring accuracy across RL checkpoints.}
At early, middle, and late stages of GRPO training, we replayed eight
LIBERO action chunks for each pair of pre- and post-update policies.
We recomputed the approximate log-ratios in FP32 and compared them
with reference values obtained using a converged inverse and an
explicit Jacobian. On these samples, the mean absolute error was
below 0.0002 at all three checkpoints.
\end{minipage}\hfill
\begin{minipage}[t]{0.35\linewidth}
\vspace{0pt}\centering\small
\setlength{\tabcolsep}{3pt}
\captionof{table}{FP32 log-ratio error across GRPO checkpoints.}
\label{tab:fp32-logratio}
\begin{tabular}{@{}lr@{}}
\toprule
\makecell[l]{RL\\checkpoint} &
\makecell[r]{Mean absolute\\log-ratio error\\(FP32)} \\
\midrule
Early  & 0.000181 \\
Middle & 0.000070 \\
Late   & 0.000061 \\
\bottomrule
\end{tabular}
\end{minipage}

\section{Implementation Details}
\label{app:impl}

\xhdr{Model}
The action expert of $\pi_{0.5}$ is extended with two additive
pathways, each a two-layer MLP of width 1024 with swish
activation and a zero-initialised output layer, so that the
extended model reproduces the pretrained velocity at
initialisation.
The interval pathway embeds $t-r$ with a sinusoidal positional
encoding (periods $4\times10^{-3}$ to $4$); the path pathway
projects the martingale increment $\Delta M_{r,t}$ through the
action input projection, mean-pools it over the horizon, adds the
interval encoding, and passes the sum through the MLP.
Both outputs enter the expert through its adaptive RMS-norm
conditioning, not as additional tokens.
The vision--language backbone (SigLIP and Gemma-2B) is frozen;
the action expert, its projections, and the two pathways are
trained, all in bfloat16 under FSDP without parameter sharding.
The expert predicts a chunk of 10 steps in a 32-dimensional padded
action space; the first 5 steps of the first 7 dimensions are
executed, and the conditional density is
evaluated on these $d=5\times7=35$ coordinates only.

\xhdr{Brownian path}
Each of the $d$ coordinates carries an independent standard
Brownian motion $B$ on $[0,1]$, represented by its first $L=5$
Karhunen--Lo\`eve modes,
$B(q)=\sum_{k=1}^{L}\phi_k(q)\,\xi_k$ with
$\phi_k(q)=\sqrt2\,\sin(\omega_k q)/\omega_k$,
$\omega_k=(k-\tfrac12)\pi$, and $\xi\sim\gN(0,I_L)$.
The diffusion profile $\beta(t)=\sqrt{2(1-t)}$ is implemented as
the time change $M_t=\sigma\bigl[B(1)-B((1-t)^2)\bigr]$, so that the
increment of \cref{eq:method_increment} is the closed-form
difference
$\Delta M_{r,t}=\sigma\bigl[B((1-r)^2)-B((1-t)^2)\bigr]$ with
$c_k(r,t)=\phi_k((1-r)^2)-\phi_k((1-t)^2)$; no quadrature is
used.
For the full jump, $\|c(0,1)\|_2^2=\sum_{k=1}^{5}2/\omega_k^2
=0.9596$, so the realized variance in \cref{eq:method_exact_density}
is $\Sigma=0.9596\,\sigma^2$ rather than its continuum limit
$\sigma^2$; the same truncation is used at training and at
scoring, since a mismatch would bias every importance ratio.

\xhdr{Distillation}
The student is initialised from the $\pi_{0.5}$ SFT checkpoint
and trained for 1{,}000 optimizer steps on 4 A100 GPUs with global
batch 128 (micro-batch 4), AdamW with $\beta=(0.9,0.95)$,
$\epsilon=10^{-8}$, weight decay $10^{-10}$, and gradient
clipping at 1.0.
The learning rate is warmed up linearly to $5\times10^{-6}$ over
the run.
The objective is $\gL_{\mathrm{Diag}}+\lambda\gL_{\mathrm{LSD}}$
with $\lambda=1$, where the consistency term is switched on after
the first 5\% of steps; the consistency residual is not
normalised and is typically one to two orders of magnitude larger
than the diagonal residual, so $\lambda=1$ does not correspond to
an equal weighting of the two terms.
Interval endpoints are drawn as $r=\mathrm{mid}-p\,\delta/2$,
$t=\mathrm{mid}+p\,\delta/2$, where $\mathrm{mid}$ and $\delta$
are the midpoint and length of the interval between two
independent $\gU[0,1)$ draws and $p$ increases linearly from 0 at
5\% of the run to 1 at 25\%; a quarter of each batch is forced to
the full interval $(r,t)=(0,1)$.
The noise scale is ramped linearly from 0 to $\sigma=0.1$ over the
first 30\% of the run and held fixed thereafter.
The checkpoint at step 600 is used.

\xhdr{Reinforcement learning}
GRPO runs in RLinf on 4 A100 GPUs.
Each iteration collects 8 groups of $N=8$ rollouts on LIBERO
(32 parallel environments, two rollout epochs) and 32 groups on
MetaWorld, with the same task and initial state within a group,
$K=1$, and $\sigma=0.1$ applied in full from the first iteration.
Episodes run for at most 500 environment steps on LIBERO and 100 on
MetaWorld; the return is 1 at the step of first success and 0
otherwise.
Groups whose mean return lies outside $[0.1,0.9]$ are discarded,
which removes degenerate groups before the update.
Advantages use $\kappa=10^{-6}$ and are normalised across the
batch; the clipped objective uses $\epsilon=0.2$ with a dual clip
at 3.0, token-mean aggregation, no entropy bonus, and no KL
penalty.
Each iteration performs one pass over the retained rollout chunks in minibatches of 256 with micro-batch 64, using AdamW with
learning rate $10^{-6}$,
$\beta=(0.9,0.95)$, weight decay 0.01, and gradient clipping at
1.0, without warmup or decay.
The Jacobian-vector products in \cref{eq:method_losses,eq:method_approx_density}
are computed by forward-mode automatic differentiation.

\textbf{Training budget and wall-clock time.}
The original It\^o + GRPO runs collected 64 rollouts per iteration on LIBERO and 256 on MT50. We report the full training budgets below.
\par\smallskip
{\centering\small
\captionof{table}{Training budgets for the original It\^o + GRPO runs.}
\label{tab:grpo-training-budget}
\begin{tabular}{@{}lrrr@{}}
\toprule
Benchmark & RL iterations & Optimizer updates & Rollouts collected \\
\midrule
LIBERO & 600 & 15,000 & 38,400 \\
MT50 & 300 & 7,500 & 76,800 \\
\bottomrule
\end{tabular}\par}
\smallskip
End-to-end RL time includes policy scoring. The It\^o times are estimates for the original 600-iteration LIBERO run and 300-iteration MT50 run.
\par\smallskip
{\centering\small
\captionof{table}{End-to-end GRPO training time, including policy scoring.}
\label{tab:grpo-training-time}
\begin{tabular}{@{}lrr@{}}
\toprule
Training method & LIBERO (hours) & MT50 (hours) \\
\midrule
It\^o + GRPO (estimated) & 33.3 & 49.3 \\
Isotropic output noise + GRPO & 74.8 & 72.1 \\
\bottomrule
\end{tabular}\par}

\xhdr{Evaluation}
LIBERO is evaluated on the four ten-task suites with 500 episodes
per suite in 500 parallel environments; MetaWorld MT50 on 50 tasks
with 10 fixed reset states each, following $\pi_{\mathrm{RL}}$.
Forward times in \cref{tab:main} are wall-clock times of the action
head per chunk on one A100.

\xhdr{Successful Rollouts}
\label{app:rollouts}
We visualize successful rollouts of the one-step \nameshort policy (It\^o map, $K{=}1$, $\sigma{=}0$ at evaluation) on the four LIBERO suites and on MetaWorld MT50. Each figure shows uniformly spaced frames from a single successful episode, from the initial state to the step at which success is registered.

\begin{figure}[htbp!]
    \centering
    \includegraphics[width=0.90\textwidth]{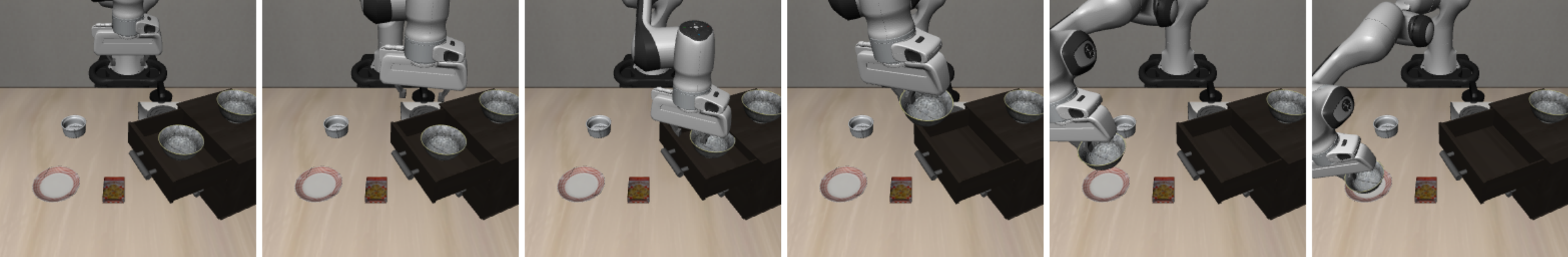}
    \caption{\texttt{LIBERO-Spatial}, task 4. Pick up the black bowl in the top drawer of the wooden cabinet and place it on the plate.}
    \label{fig:rollout-libero-spatial-t4}
\end{figure}
\looseness=-1

\begin{figure}[htbp!]
    \centering
    \includegraphics[width=0.90\textwidth]{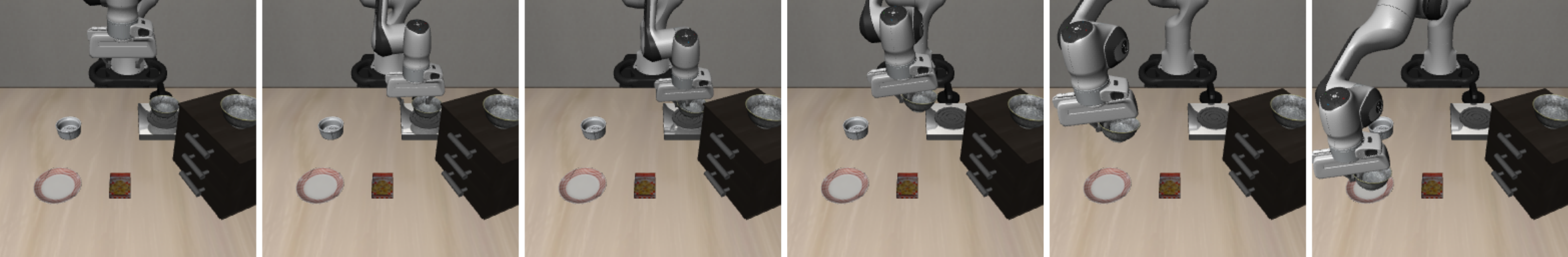}
    \caption{\texttt{LIBERO-Spatial}, task 7. Pick up the black bowl on the stove and place it on the plate.}
    \label{fig:rollout-libero-spatial-t7}
\end{figure}
\looseness=-1

\begin{figure}[htbp!]
    \centering
    \includegraphics[width=0.90\textwidth]{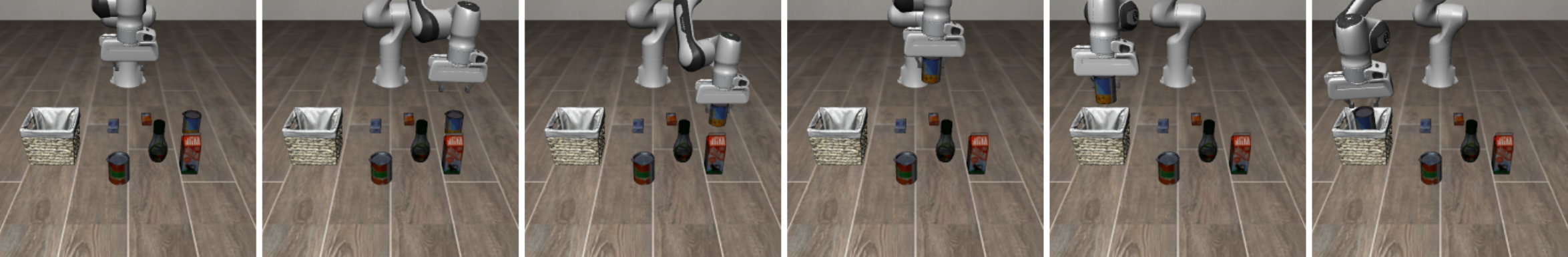}
    \caption{\texttt{LIBERO-Object}, task 0. Pick up the alphabet soup and place it in the basket.}
    \label{fig:rollout-libero-object-t0}
\end{figure}
\looseness=-1

\begin{figure}[htbp!]
    \centering
    \includegraphics[width=0.90\textwidth]{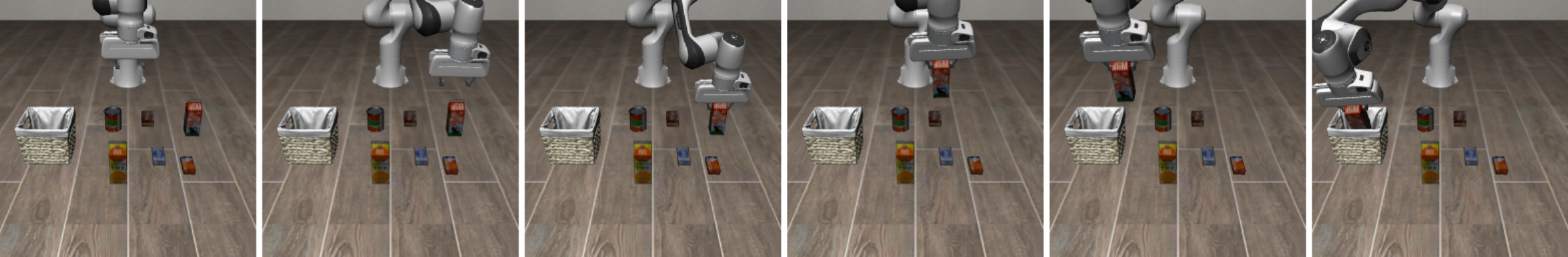}
    \caption{\texttt{LIBERO-Object}, task 7. Pick up the milk and place it in the basket.}
    \label{fig:rollout-libero-object-t7}
\end{figure}
\looseness=-1

\begin{figure}[htbp!]
    \centering
    \includegraphics[width=0.90\textwidth]{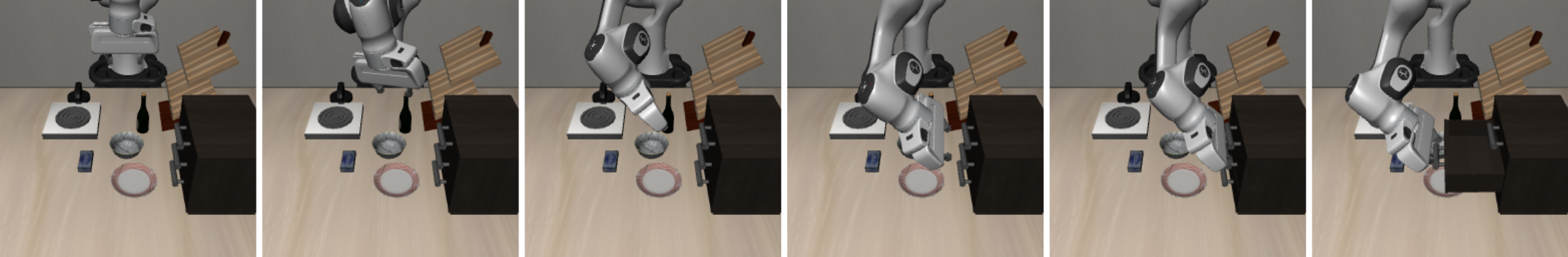}
    \caption{\texttt{LIBERO-Goal}, task 0. Open the middle drawer of the cabinet.}
    \label{fig:rollout-libero-goal-t0}
\end{figure}
\looseness=-1

\begin{figure}[htbp!]
    \centering
    \includegraphics[width=0.90\textwidth]{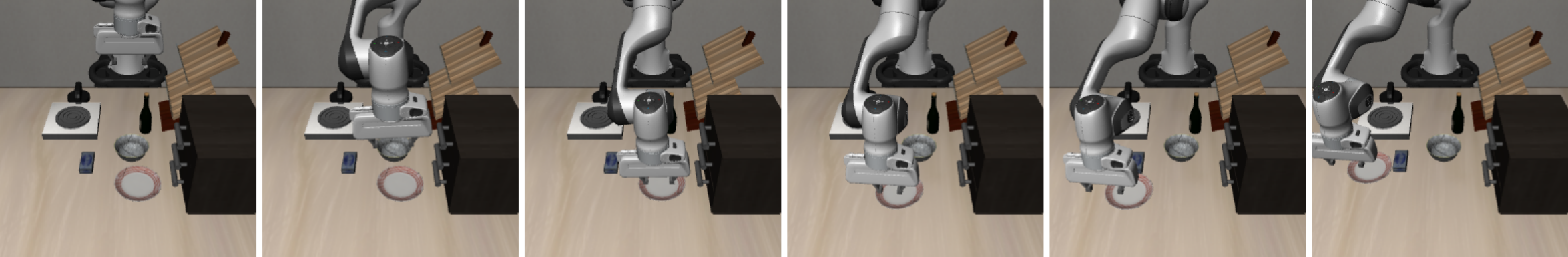}
    \caption{\texttt{LIBERO-Goal}, task 5. Push the plate to the front of the stove.}
    \label{fig:rollout-libero-goal-t5}
\end{figure}
\looseness=-1

\begin{figure}[htbp!]
    \centering
    \includegraphics[width=0.90\textwidth]{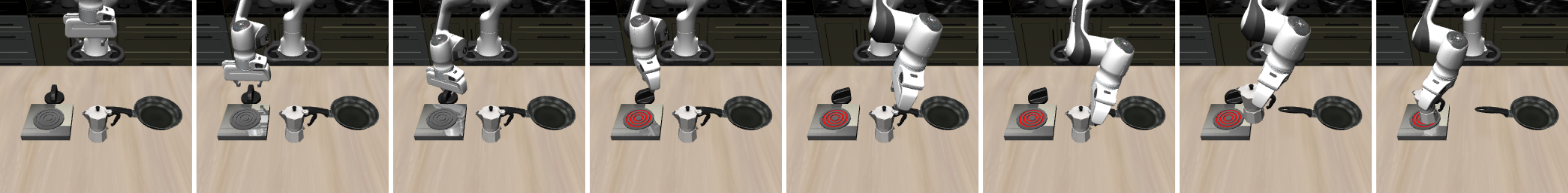}
    \caption{\texttt{LIBERO-Long}, task 2. Turn on the stove and put the moka pot on it.}
    \label{fig:rollout-libero-long-t2}
\end{figure}
\looseness=-1

\begin{figure}[htbp!]
    \centering
    \includegraphics[width=0.90\textwidth]{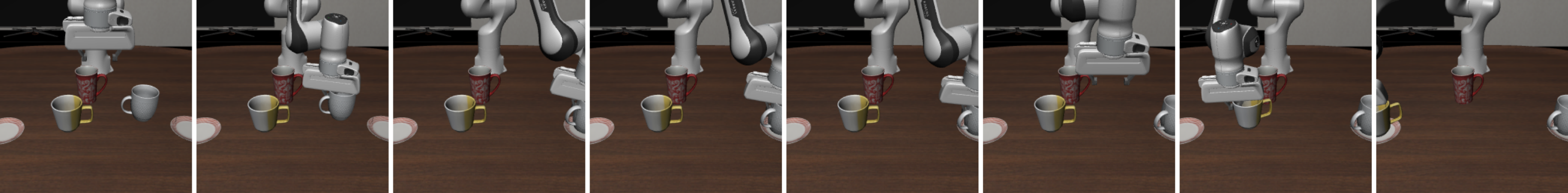}
    \caption{\texttt{LIBERO-Long}, task 4. Put the white mug on the left plate and put the yellow and white mug on the right plate.}
    \label{fig:rollout-libero-long-t4}
\end{figure}
\looseness=-1

\begin{figure}[htbp!]
    \centering
    \includegraphics[width=0.90\textwidth]{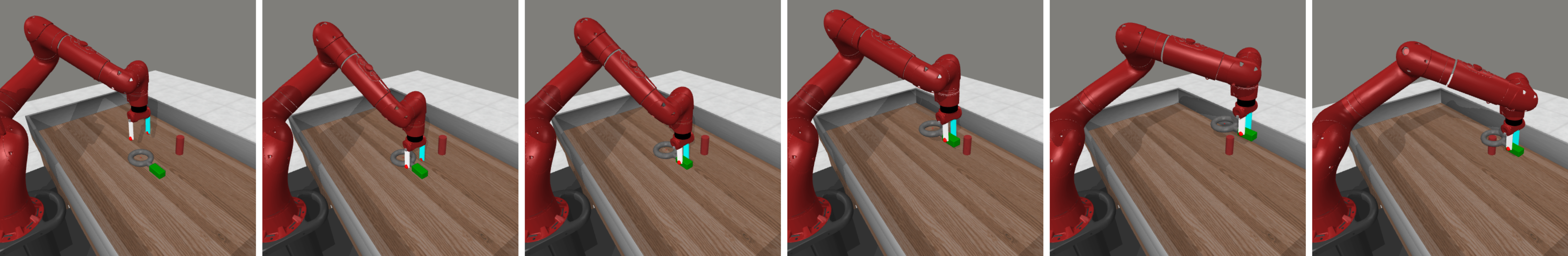}
    \caption{\texttt{assembly-v3} (MetaWorld MT50). Pick up a nut and place it onto a peg.}
    \label{fig:rollout-metaworld-assembly}
\end{figure}
\looseness=-1

\begin{figure}[htbp!]
    \centering
    \includegraphics[width=0.90\textwidth]{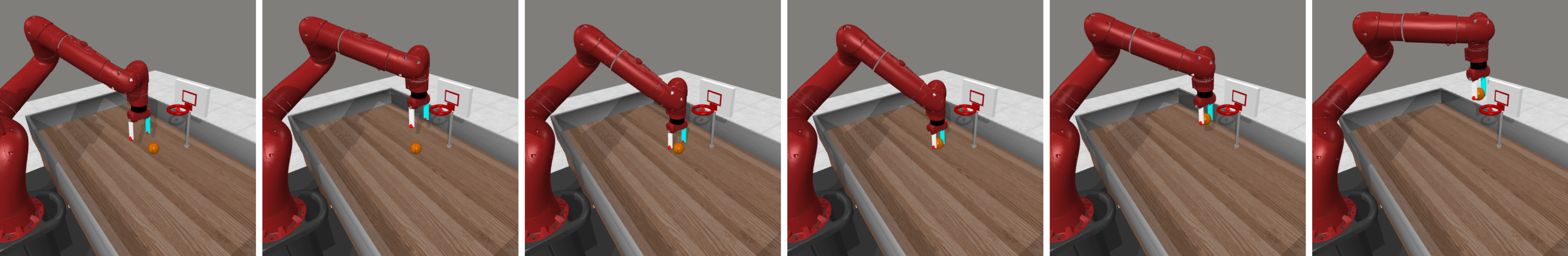}
    \caption{\texttt{basketball-v3} (MetaWorld MT50). Dunk the basketball into the basket.}
    \label{fig:rollout-metaworld-basketball}
\end{figure}
\looseness=-1

\begin{figure}[htbp!]
    \centering
    \includegraphics[width=0.90\textwidth]{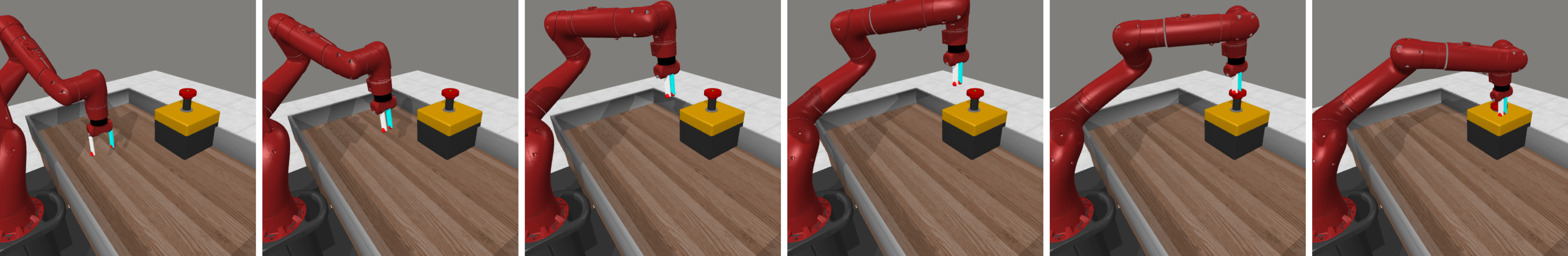}
    \caption{\texttt{button-press-topdown-v3} (MetaWorld MT50). Press a button from the top.}
    \label{fig:rollout-metaworld-button-press-topdown}
\end{figure}
\looseness=-1

\begin{figure}[htbp!]
    \centering
    \includegraphics[width=0.90\textwidth]{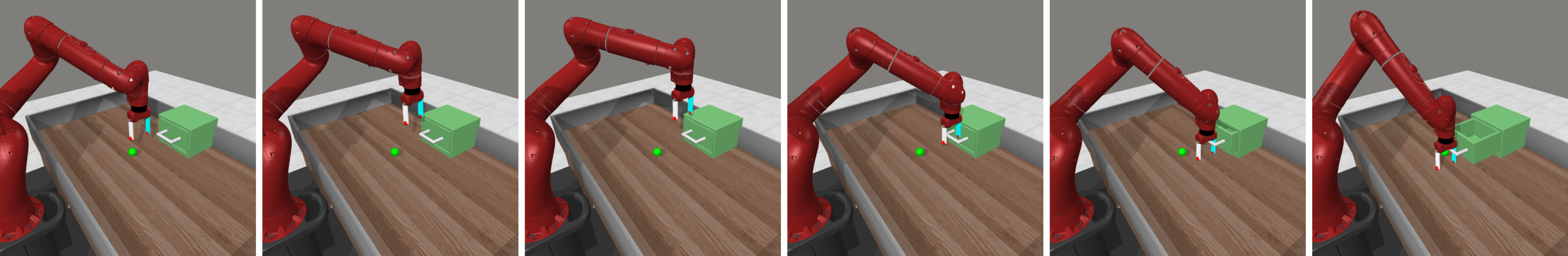}
    \caption{\texttt{drawer-open-v3} (MetaWorld MT50). Open a drawer.}
    \label{fig:rollout-metaworld-drawer-open}
\end{figure}
\looseness=-1

\begin{figure}[htbp!]
    \centering
    \includegraphics[width=0.90\textwidth]{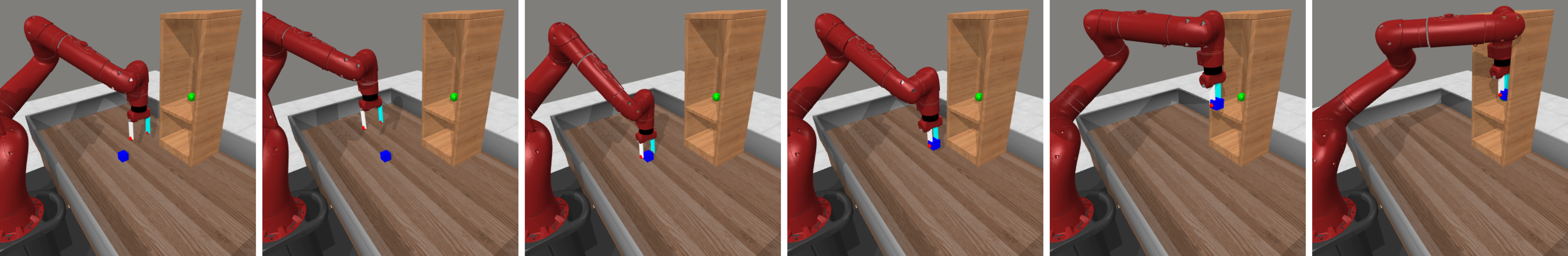}
    \caption{\texttt{shelf-place-v3} (MetaWorld MT50). Pick and place a puck onto a shelf.}
    \label{fig:rollout-metaworld-shelf-place}
\end{figure}
\looseness=-1

\begin{figure}[htbp!]
    \centering
    \includegraphics[width=0.90\textwidth]{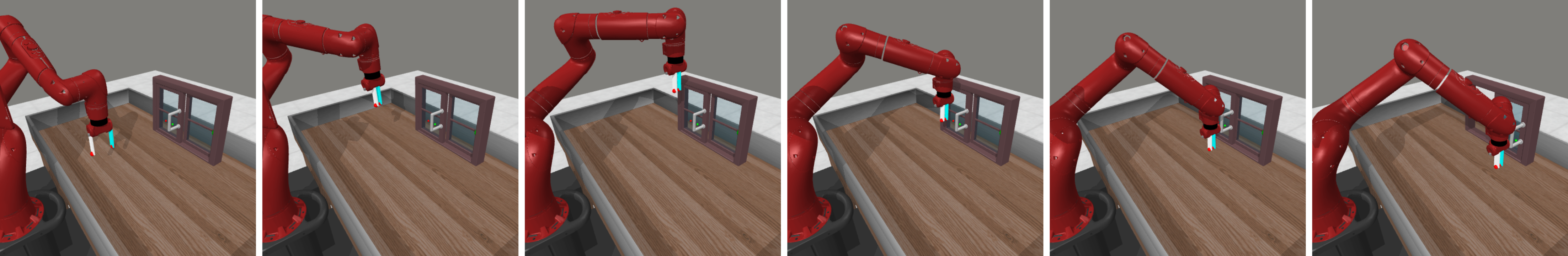}
    \caption{\texttt{window-open-v3} (MetaWorld MT50). Push and open a window.}
    \label{fig:rollout-metaworld-window-open}
\end{figure}
\looseness=-1

\section{Effect of Self-Distillation Before GRPO}
\label{app:directgrpo}

Both LIBERO runs started from the same $\pi_{0.5}$ SFT checkpoint. We applied self-distillation before GRPO in one run and trained the other directly with one-step It\^o + GRPO. We kept the GRPO settings fixed and evaluated both policies with one-step sampling and exploration noise turned off. The distilled policy started from a higher success rate and retained a 21.45-percentage-point lead after 100 optimizer updates.
\par\smallskip
{\centering\small
\begin{tabular}{@{}lrr@{}}
\toprule
& \multicolumn{2}{c}{LIBERO success (\%)} \\
\cmidrule(l){2-3}
Training pipeline & Before RL & After 100 updates \\
\midrule
Self-distillation followed by one-step GRPO & 68.60 & 73.50 \\
Direct one-step GRPO & 40.55 & 52.05 \\
\bottomrule
\end{tabular}\par}

\section{Comparison with Four-Step ReinFlow-Style PPO}
\label{app:reinflow}

On each benchmark, both methods started from the same $\pi_{0.5}$ SFT checkpoint before method-specific training. We evaluated both policies on the same initial states after 100 optimizer updates. At this checkpoint, \nameshort led by 32.95 percentage points on LIBERO and 4.00 points on MT50.
\par\smallskip
{\centering\small
\begin{tabular}{@{}llrrr@{}}
\toprule
Benchmark & Method & Passes & Success (\%) & Head time (ms/chunk) \\
\midrule
LIBERO & \nameshort + GRPO & 1 & 73.50 & 7.6 \\
       & ReinFlow-style PPO & 4 & 40.55 & 68.5 \\
\midrule
MT50   & \nameshort + GRPO & 1 & 59.40 & 7.6 \\
       & ReinFlow-style PPO & 4 & 55.40 & 80.0 \\
\bottomrule
\end{tabular}\par}

% \bibliography{iclr2027_conference}
% \bibliographystyle{iclr2027_conference}

% \appendix
% \section{Appendix}
% \input{contents/08_appendix}

\end{document}